\documentclass[letterpaper, 10 pt, conference]{ieeeconf}  

\IEEEoverridecommandlockouts                              

\usepackage{graphics} 
\usepackage{epsfig} 
\usepackage{amsmath} 
\usepackage{amssymb}  
\usepackage{tabularx} 
\usepackage{etoolbox}

\usepackage{preambles}

\title{\LARGE \bf
Model-Informed Safe Reinforcement Learning for Bipedal Locomotion via Step-to-Step Prediction
}

\author{Victor Paredes$^{1}$ and Ayonga Hereid$^{1}$
\thanks{This work was supported in part by the National Science Foundation under grant FRR-21441568. }
\thanks{$^{1}$Mechanical and Aerospace Engineering,
        The Ohio State University, Columbus, OH, USA.
        {\tt\small \{paredescauna.1,hereid.\}@osu.edu}}%
}

\begin{document}

\maketitle
\thispagestyle{empty}
\pagestyle{empty}


\begin{abstract}

Humanoid robots promise versatile mobility in cluttered, human-centric environments, but real deployment demands principled safety. Classical model-based gait generators yield interpretable motions but often lack the robustness and adaptability of modern reinforcement learning (RL) based approaches. We propose a model-informed reinforcement learning framework anchored to the analytical Angular Momentum Linear Inverted Pendulum (ALIP) template. We provide a step-to-step safety certificate for ALIP stepping via a discrete exponential control barrier function (DECBF) and use it as (i) a training-time shaping signal and (ii) a runtime action filter that minimally adjusts swing-foot placement to satisfy template-level constraints. Full-order safety is evaluated empirically on the Digit humanoid in MuJoCo with a whole-body controller stack. Compared to an unconstrained baseline, our approach reduces safety-violation events in the reported external-disturbance trial, while larger lateral-velocity transients reveal a safety--tracking tradeoff.

\end{abstract}


\section{Introduction}
\label{sec:introduction}
Humanoid locomotion enables agile mobility in cluttered, human-centric environments, but real-world deployment is inherently safety-critical. In legged systems, safety must be assessed over a predictive horizon: foot placement decisions made at the current step directly determine the robot’s state at the next impact event. Ensuring that these step-to-step transitions remain within safe limits is therefore essential for reliable walking.
Classical model-based gait generators provide interpretable, predictive structure by leveraging reduced-order dynamics and hybrid system analysis. However, while these approaches enable structured reasoning about stability, they often struggle to generalize across varying terrains, gait modes, or disturbances without extensive retuning. In contrast, reinforcement learning (RL) methods have demonstrated remarkable robustness and adaptability in complex environments, yet they typically lack explicit step-to-step predictive structure or mathematical safety guarantees. As a result, current RL-based locomotion policies remain difficult to interpret or certify at the impact horizon where safety is most critical.


\begin{figure}[t]
\centering
\vspace{2mm}
\includegraphics[width=\linewidth]{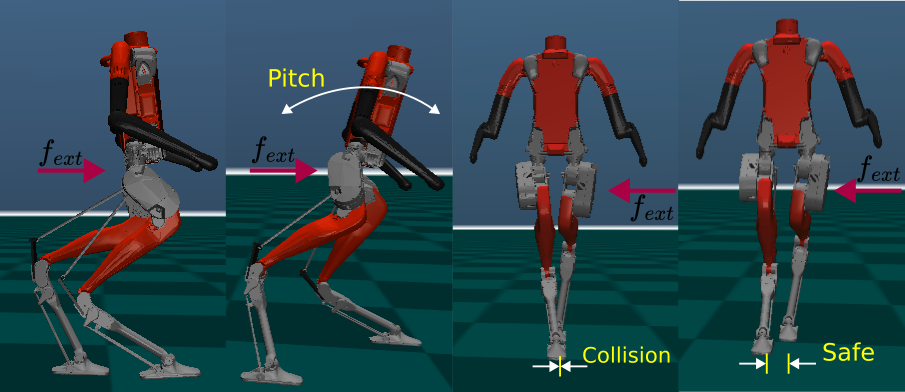}
\caption{Overview of the ALIP-informed safe RL framework on the Digit humanoid. The policy commands swing-foot placement and torso pitch. DECBFs derived from the ALIP model constrain foot-placement actions during training (safety shaping) and execution (projection filter).}  
\label{fig:DigitMoving}
\end{figure}

\begin{figure*}[t]
\centering
\vspace{2mm}
\includegraphics[width=0.90\linewidth]{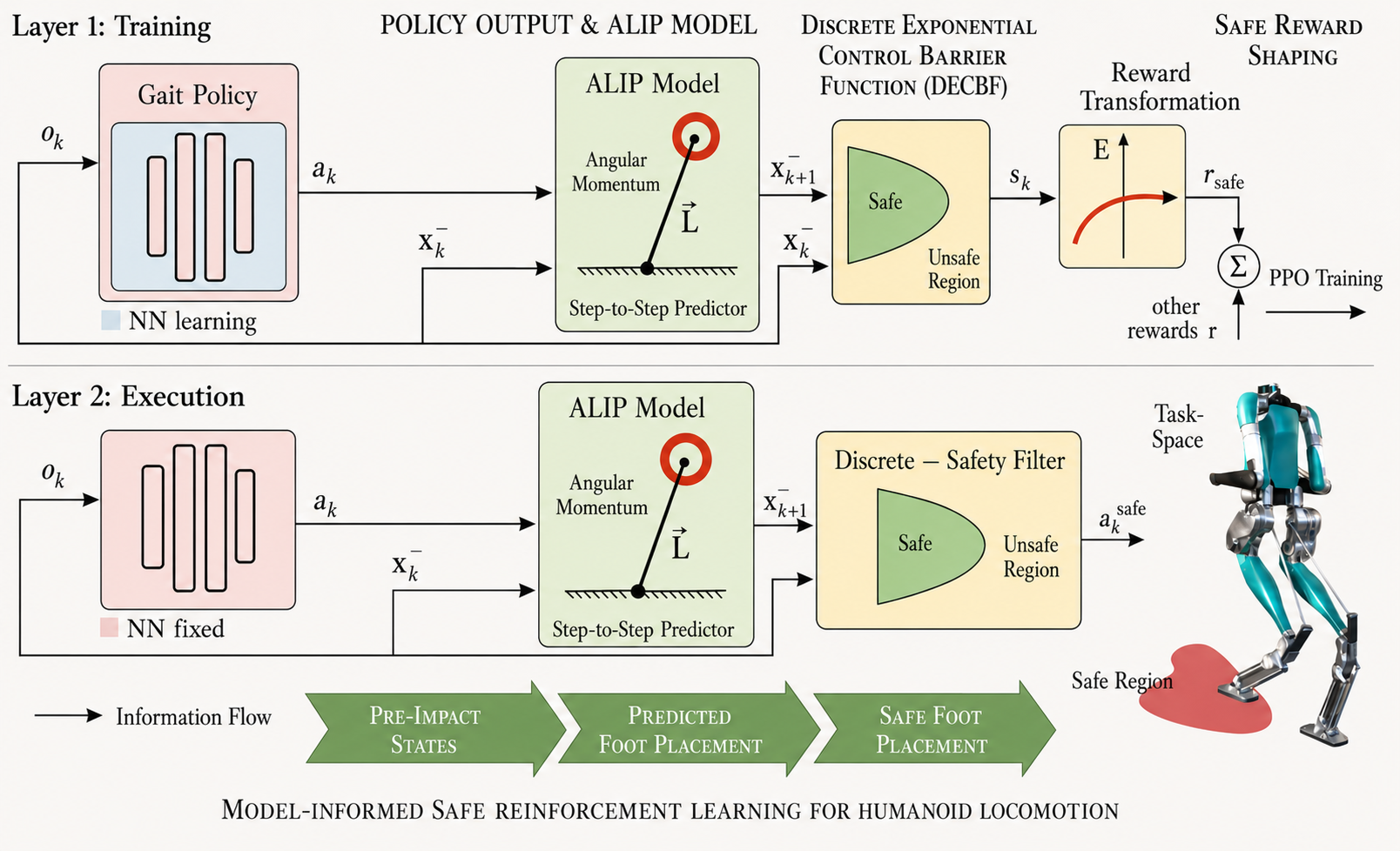}
\caption{Training vs.\ execution pipeline. During training, the policy outputs action $\mathbf{a}_k=(u_k,\theta_k)$; the foot-placement component $u_k$ is evaluated through the ALIP step-to-step map to compute the safety certificate $\mathbf{s}_k$, yielding a shaping reward $r_{\text{safe}}$. During execution, the shaping reward is replaced by a projection that finds the closest $u_k$ satisfying $\mathbf{s}_k \geq 0$ before passing the command to the whole-body controller.}
\label{fig: SafeRL_Diagram}
\end{figure*}

Model-based techniques have historically formed the foundation of humanoid locomotion~\cite{yamamoto2020survey, tong2024advancements}, ranging from full-order approaches such as Hybrid Zero Dynamics (HZD)~\cite{westervelt2003hybrid, sreenath2011compliant, hereid2015hybrid, paredes2020dynamic} to simplified models like the Linear Inverted Pendulum (LIP)~\cite{kajita2010biped}, the spring-loaded LIP (SLIP)~\cite{xiong2020dynamic}, and other reduced-order models~\cite{brown2016reaction,liu2016terrain}. Among simplified models, the Hybrid-LIP (H-LIP)~\cite{xiong2021slip, paredes2022resolved, xiong20223} and the Angular Momentum LIP (ALIP)~\cite{gong2022zero, gibson2022terrain, gao2023time} capture step-to-step dynamics that map the robot's state from one impact to the next. The ALIP, in particular, uses contact angular momentum instead of linear Center of Mass (CoM) velocity, yielding a more accurate predictor with simple closed-form expressions that are highly interpretable.

On the model-free side, Reinforcement Learning (RL) has demonstrated state-of-the-art robustness and agility~\cite{garcia2020teaching, castillo2020hybrid, li2021reinforcement, luo2023robust, castillo2024data, radosavovic2024real} by implicitly learning complex dynamics through extensive training in physics-based simulations. End-to-end policies that output joint-level references are effective but difficult to interpret or certify at the step-to-step horizon where foot placement is decisive. Policies that instead generate task-space commands, such as desired swing-foot positions, offer greater interpretability and are more amenable to structured safety analysis.

Despite this task-space abstraction, current RL policies lack a mathematical model that predicts the next impact state as a function of foot placement, so safety at the impact horizon cannot be enforced through structured constraints. The key idea of this work is to embed the ALIP step-to-step predictor into the RL pipeline and construct a discrete-time exponential control barrier function (DECBF)~\cite{ames2019control} on the resulting map. The DECBF is used (i)~as safety shaping that penalizes actions whose predicted next state would exit a safe set, and (ii)~as a projection-based safety filter at execution that minimally adjusts foot placement. Unlike continuous-time ECBFs applied within QP-based whole-body control~\cite{mistry2010inverse, herzog2016momentum, reher2021inverse, nguyen20163d, paredes2024safe}, our discrete-time formulation operates at the planning layer, where the one-dimensional feasible sets admit closed-form projection for fixed step time. Full-order safety under the Digit controller stack is evaluated empirically.

We make three contributions: (1) derive an affine-in-foot-placement discrete-time safety condition (DECBF) for ALIP stepping; (2) use this condition as a training-time shaping term and a runtime projection filter; and (3) demonstrate reduced safety-violation events in Digit MuJoCo simulation under the reported periodic disturbance scenario.

\section{Angular-Momentum-Based Dynamics}
\label{sec:alip-rl}
This section reviews the Angular-Momentum Linear Inverted Pendulum (ALIP) dynamics~\cite{gong2022zero,gibson2022terrain,gao2023time} and derives a step-to-step map in which foot placement $(u_{x,k},u_{y,k})$ serves as the control input.
Consider single-contact walking with horizontal template states
\[
\xstate=\begin{bmatrix} p_x \\ L_y \end{bmatrix},\qquad
\ystate=\begin{bmatrix} p_y \\ L_x \end{bmatrix},
\]
where $p_\square$ is the CoM position in the stance frame and $L_\square$ is the corresponding \emph{contact} angular momentum component. With a control-enforced constant CoM height $H$ and mass $m$, the continuous-time dynamics take the ALIP form:
\begin{align}
    \dot{\xstate} =
    \underbrace{\begin{bmatrix} 0 & \frac{1}{mH} \\[2pt] m g & 0 \end{bmatrix}}_{\bar A_x}\xstate,\quad
    \dot{\ystate} =
    \underbrace{\begin{bmatrix} 0 & -\frac{1}{mH} \\[2pt] -m g & 0 \end{bmatrix}}_{\bar A_y}\ystate.\label{eq:alip-ct}
\end{align}
The ALIP template is a reduced-order approximation of full-order humanoid dynamics; it assumes fixed stance contact, constant CoM height, and negligible centroidal angular momentum about the CoM during single support. We use the nominal predictor for safety shaping and filtering, while deviations are treated as model mismatch.
Let $\ell \doteq \sqrt{g/H}$ and $T$ be a fixed step duration, and define $s_T\doteq\sinh(\ell T)$ and $c_T\doteq\cosh(\ell T)$. The nominal template prediction from time $t_0$ to $t_0{+}T$ is
\begin{align}
\xstate(t_0{+}T) &= e^{\bar A_x T}\,\xstate(t_0),\nonumber\\
\ystate(t_0{+}T) &= e^{\bar A_y T}\,\ystate(t_0),\label{eq:alip-predict}
\end{align}
where $e^{(\cdot)}$ denotes the matrix exponential.



\noindent\textbf{Impact map and step-to-step dynamics.}
We denote pre-/post-impact states by $(\cdot)_k^-$ and $(\cdot)_k^+$, respectively. Under the idealized impact map, the contact angular momentum component is impact-invariant. Let $(u_{x,k},u_{y,k})$ denote the touchdown location of the swing foot \emph{relative to the CoM} at impact, expressed in the pre-impact stance frame. Re-expressing kinematics in the new stance frame,
\begin{equation}
\xstate_k^+ = \begin{bmatrix} -u_{x,k} \\ L_{y,k}^- \end{bmatrix},\qquad
\ystate_k^+ = \begin{bmatrix} -u_{y,k} \\ L_{x,k}^- \end{bmatrix}.
\label{eq:impact}
\end{equation}
Applying \eqref{eq:alip-predict} over one step with $T_k\!\to\!T_{k+1}$ and eliminating $\xstate_k^+$ yields the discrete, affine step map (sagittal):
\begin{align}
\xstate_{k+1}^- &= \begin{bmatrix}
0 & \frac{s_T}{mH \ell} \\
0 & c_T
\end{bmatrix}\,\xstate_k^- -\begin{bmatrix}
c_T \\[1pt] mH \ell s_T
\end{bmatrix}\,u_{x,k}, \nonumber
\end{align}
and (frontal):
\begin{align}
\ystate_{k+1}^- &= \begin{bmatrix}
0 & -\frac{s_T}{mH \ell} \\
0 & c_T
\end{bmatrix}\,\ystate_k^- - \begin{bmatrix}
c_T \\[1pt] -mH \ell s_T
\end{bmatrix}\,u_{y,k}. \nonumber
\end{align}


\noindent\textbf{Orbital energy.}
Beyond the step-to-step state, each swing phase can be characterized by a conserved quantity. Pendulum models governed by~(\ref{eq:alip-predict}) exhibit a time-invariant \emph{orbital energy} that characterizes the CoM motion given an initial position and velocity:
\begin{equation}
    E\!\left([p,L]^\top\right) \;\doteq\; -\frac{g}{2H}p^2 + \frac{1}{2m^2H^2}L^2,
\end{equation}
where $[p,L]^\top$ denotes either $\xstate=[p_x,L_y]^\top$ or $\ystate=[p_y,L_x]^\top$.

\begin{figure*}[t]
\centering
\vspace{2mm}
\includegraphics[width=0.98\linewidth]{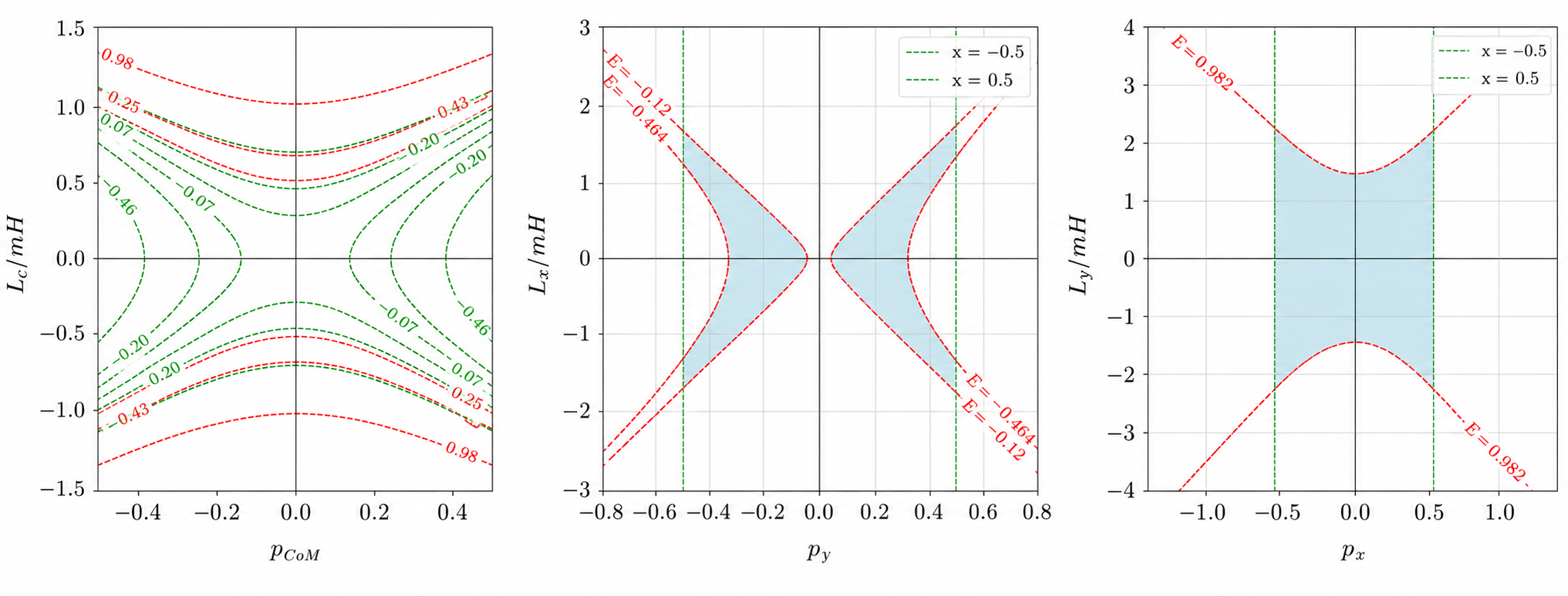}
\vspace{-1em}
\caption{Orbital energy and the orbital-energy safe regions used in this work. Left: orbital-energy phase portrait with iso-energetic lines. Middle: lateral-walking safe region ($p_y \in [-0.5, 0.5]$~m and $E \in [-0.464, -0.012]$). Right: sagittal safe region ($p_x \in [-0.7, 0.7]$~m and $E < 1.125$).}
\label{fig: OE}
\end{figure*}

The sign of $E$ determines the qualitative behavior of the passive pendulum in that plane (see Fig.~\ref{fig: OE}, left).
The \emph{apex} $(p,L)=(0,0)$ corresponds to the CoM directly above the stance foot ($p=0$) with zero velocity and is an unstable equilibrium of the ALIP.
The separatrix $E=0$ forms the capture boundary: along its stable branch, the CoM asymptotically comes to rest at the apex, which motivates the \emph{instantaneous capture point}.
When $E>0$, the system possesses sufficient energy to pass through the apex ($p$ crosses zero).
When $E<0$, the apex is unreachable and the CoM turns around before reaching $p=0$ (i.e., $p$ does not change sign).
Given any initial state $(p,L)$, the orbital energy is maintained constant along the passive trajectory, so each unforced pendulum orbit is uniquely characterized by its energy level.

\noindent\textbf{Orbital energy and leg crossing.}
For lateral walking, a gait with $E(\ystate)<0$ keeps the CoM on one side of the stance foot. If an external disturbance drives $E(\ystate)$ positive, the CoM can cross the stance foot laterally, forcing a leg-crossing motion that leads to self-collision. This observation motivates the orbital-energy safety constraint developed in Sec.~\ref{sec:alip-safe}. Combining the impact reset \eqref{eq:impact} with the within-step transition yields a closed-form orbital-energy update (neglecting the residual disturbance):
\begin{align}
E(\xstate_{k+1}^-) &= \frac{(L_{y,k}^-)^2}{2m^2H^2} - \frac{g}{2H}\,u_{x,k}^2,\\
E(\ystate_{k+1}^-) &= \frac{(L_{x,k}^-)^2}{2m^2H^2} - \frac{g}{2H}\,u_{y,k}^2.
\label{eq:Eupdate}
\end{align}

\noindent\textbf{Key structural property.}
Two features of the ALIP step-to-step map are central to our framework. First, the predicted pre-impact state $\xstate_{k+1}^-$ (and $\ystate_{k+1}^-$) is \emph{affine} in the foot-placement input $(u_{x,k},u_{y,k})$ for fixed step time $T$. Second, the orbital-energy update~\eqref{eq:Eupdate} is a known \emph{closed-form} function of foot placement and the current angular momentum. Together, these properties mean that state and separation bounds yield intervals or half-spaces in foot placement, whereas an energy envelope yields a union of at most two intervals; these one-dimensional sets admit exact nearest-point projection. In the following sections, we exploit this structure to embed safety into a reinforcement-learning pipeline: the foot-placement variables become the policy's action, and the ALIP predictor serves as the bridge that connects each action to a verifiable safety certificate at the next impact.


\section{Model-Informed Safe RL Using ALIP}
\label{sec:safe-rl}
We now embed the ALIP step-to-step map into a reinforcement-learning pipeline for humanoid locomotion. The RL policy outputs a task-space action whose primary components are the swing-foot touchdown location $(u_{x,k},u_{y,k})$ relative to the CoM (see Sec.~\ref{sec:digit-impl} for the full action and observation specification). Because the ALIP predictor maps each foot-placement action to the next pre-impact state in closed form, we can evaluate the safety implications of every candidate action \emph{before} it is executed. This enables safety enforcement at two levels: (i) \emph{during learning}, via a safety-shaping reward that penalizes actions whose \emph{predicted} next state exits a safe set, and (ii) \emph{at execution}, via a projection-based safety filter that minimally modifies the policy output to satisfy a discrete-time control barrier condition.

\subsection{ALIP Predictors and Shorthand}
From Sec.~\ref{sec:alip-rl}, with fixed step time $T$ the sagittal predictors can be written compactly as
\begin{align}
\label{eq:alip-predictors}
    p_{x,k+1}^- &= a(T)\,L_{y,k}^- - c(T)\,u_{x,k},\\
    E_{x,k+1} &= \frac{(L_{y,k}^-)^2}{2m^2H^2} - \frac{g}{2H}u_{x,k}^2,
\end{align}
with
\begin{equation}
\label{eq:a-c}
    a(T) \doteq \frac{\sinh(\ell T)}{mH\,\ell},\qquad c(T) \doteq \cosh(\ell T),\qquad \ell \doteq \sqrt{\tfrac{g}{H}}.\nonumber
\end{equation}
We use the shorthand $a\equiv a(T)$ and $c\equiv c(T)$ whenever $T$ is clear. The lateral-plane construction is analogous and is enforced independently per plane.

\subsection{Safe Set and Discrete Exponential CBF}
Let $\boldsymbol{z} \in\mathbb{R}^2$ denote a generic plane-wise template state (i.e., $\boldsymbol{z}=\xstate$ for sagittal or $\boldsymbol{z}=\ystate$ for frontal) and let the safe set be
\begin{equation}
\mathcal{S} \;=\; \{\,\boldsymbol{z}_k \;\mid\; B_i(\boldsymbol{z}_k)\ge 0,\;\forall i\in\mathcal{I}\,\}.
\label{eq:safe-set}
\end{equation}
where $\mathcal{I}$ denotes the set of safe conditions. 
Given a predictor that maps $(\xstate_k^-,u_k)$ to a predicted $\xstate_{k+1}^-$, the discrete exponential CBF (DECBF) condition is
\begin{equation}
\Delta B_i(\xstate_k^-,u_k,T) + \gamma\,B_i(\xstate_k^-) \;\ge\; 0,\qquad \forall i \in \mathcal{I},
\label{eq:decbf}
\end{equation}
where $\Delta B_i(\xstate_k^-,u_k,T) \doteq B_i(\xstate_{k+1}^-) - B_i(\xstate_k^-)$ and $\gamma\in(0,1)$.
If $B_i(\xstate_0^-)\ge 0$ and \eqref{eq:decbf} holds for all $k$, then $\xstate_k^-\in\mathcal{S}$ for all $k$ (forward invariance).
We introduce a single auxiliary variable to represent the safety certificate:
\begin{equation}
    s_i(\xstate_k^-,u_k,T) \;\doteq\; \Delta B_i(\xstate_k^-,u_k,T) + \gamma\,B_i(\xstate_k^-),
\end{equation}
which facilitates the construction of a safety function for our RL pipeline.
Note that $s_i$ depends on $u_k$ through the predicted next state $\xstate_{k+1}^-$ obtained from the ALIP step-to-step map.

\noindent\textbf{Scope.} The DECBF condition certifies forward invariance of the template safe set $\mathcal{S}$ for the reduced-order ALIP step-to-step map used in prediction. In this paper, $u_k$ denotes the CoM-relative foot-placement action (e.g., $u_k=[u_{x,k},u_{y,k}]^\top$ at touchdown); any additional action components (e.g., torso pitch) are passed through unfiltered. Because the full robot may deviate from the template due to model mismatch and imperfect low-level tracking, full-order safety is evaluated empirically under the Digit estimator and whole-body controller stack.



\subsection{How prediction is used}
\textbf{Learning-time shaping.} For a candidate action $u_k$, we evaluate $s_i(\xstate_k^-,u_k,T)$ and augment the reward with
\begin{equation}
r_{\text{safe}} = -\sum_{i\in\mathcal{I}} \eta_i \left(e^{-k_s s_i}-1\right)\,\mathbf{1}\{s_i<0\},
\label{eq:rsafe}
\end{equation}
Here $\eta_i > 0$ is a per-constraint weight and $k_s > 0$ controls the penalty steepness. This term penalizes actions predicted to violate safety, thereby steering exploration toward safer behaviors.

\textbf{Execution-time projection filter.} During execution, given a nominal foot-placement action $u_k^{\text{nom}}$, we project to the nearest action satisfying the DECBF constraints:
\begin{equation}
\label{eq:proj-filter}
u_k^{*} \;=\; \arg\min_{u}\;\|u-u_k^{\text{nom}}\|^2
\quad \text{s.t.}\quad s_i(\xstate_k^-, u, T)\ge 0,\;\forall i\in\mathcal{I}.
\end{equation}
Because the ALIP predictor makes $\xstate_{k+1}^-$ affine in $u$ for fixed $T$, reach and separation bounds yield intervals or half-spaces, while quadratic energy bounds can produce a union of at most two intervals. The specific barrier functions and the resulting constraint geometry are developed in Sec.~\ref{sec:alip-safe}; in each case the projection~\eqref{eq:proj-filter} reduces to snapping $u_k^{\text{nom}}$ to the nearest point in the one-dimensional feasible set.


\section{ALIP-Informed Safety Constraints}
\label{sec:alip-safe}
We now instantiate the generic DECBF framework (Sec.~\ref{sec:safe-rl}) with three barrier functions that address distinct safety concerns for bipedal walking: (i)~lateral separation to prevent leg crossing at touchdown, (ii)~an orbital-energy envelope to prevent the CoM from crossing the stance foot mid-swing, and (iii)~CoM-reach bounds to avoid over-extension. Each barrier is defined on the ALIP step-to-step map and yields a one-dimensional feasible set that can be projected in closed form for fixed step time.

\subsection{Lateral separation (no leg crossing)}
Let $\sigma_k\!\in\!\{+1,-1\}$ denote the support side at step $k$ (right support: $+1$, left support: $-1$). Let $d_{y,k}$ denote the signed lateral distance from stance foot to swing foot at touchdown (positive during right support and negative during left support). We impose the minimum-separation constraint
\begin{equation}
    B_{\text{sep}}(d_{y,k}) = \sigma_k d_{y,k} - W_{\min} \geq 0,
    \label{eq:sep-barrier}
\end{equation}
With the CoM-relative touchdown action, $d_{y,k}=p_{y,k}^-+u_{y,k}$ in the pre-impact stance frame, so \eqref{eq:sep-barrier} is affine in $u_{y,k}$.

\subsection{Orbital-energy envelope (leg-crossing avoidance)}
We employ the plane-wise orbital energy $E(\cdot)$ from Sec.~\ref{sec:alip-rl} evaluated on the lateral ALIP state $\ystate=[p_y,L_x]^\top$. In this plane, $E(\ystate)>0$ indicates the CoM can pass through the apex ($p_y$ crosses zero), which tends to induce leg crossing. We therefore enforce an upper bound via
\begin{equation}
    B_E(\ystate^-_k) = -E(\ystate_k^-) + E^{max}_{ALIP} \geq 0
    \label{eq: P1-Orbital-Energy}
\end{equation}
where $E^{max}_{ALIP}$ is the maximum energy level desired (chosen negative for lateral walking). In our lateral-walking experiments, we also bound the lateral CoM displacement (e.g., $p_y \in [-0.5,0.5]$~m), yielding the safe region shown in Fig.~\ref{fig: OE} (middle).

\subsection{CoM Reach (Stance-Relative Extension)}
To avoid over-leaning/unreachable steps, bound the next-step CoM position relative to stance:
\begin{equation}
p_{\min} \;\le\; p_{x,k+1}^- \;\le\; p_{\max}. \label{eq:reach-bounds}
\end{equation}
Note that~\eqref{eq:reach-bounds} yields two linear inequalities in $u_{x,k}$. We define the corresponding barrier functions as
\begin{align}
\tilde B_{x,\min} &= p_{x,k+1}^- - p_{\min} \;\ge 0,\\
\tilde B_{x,\max} &= p_{\max} - p_{x,k+1}^- \;\ge 0,
\end{align}
and apply the DECBF condition~\eqref{eq:decbf} to the reward function of the form~\eqref{eq:rsafe}.

Furthermore, the maximum average speed in the $x$-direction can be constrained by limiting the sagittal orbital energy. Using $L_y = mH v_x$, the ALIP energy satisfies $E = -\frac{g}{2H}p_x^2 + \frac{1}{2}v_x^2$; thus, we choose an energy bound corresponding to a desired speed cap (e.g., $E_{\max}=\frac{1}{2}v_{x,\max}^2$ at $p_x=0$). When combined with the robot’s mechanical limits on CoM extension, $p_{\text{min}} = -0.7$~m and $p_{\text{max}} = 0.7$~m, this bound defines the safe region used in the planning framework, as shown in Fig.~\ref{fig: OE} (right).

\subsection{Projection Geometry}
Although the energy constraint need not be globally convex, the projection filter~\eqref{eq:proj-filter} admits an efficient closed-form solution. The CoM-reach bounds~\eqref{eq:reach-bounds} together with the sagittal predictor $p_{x,k+1}^- = a\,L_{y,k}^- - c\,u_{x,k}$ imply the interval $u_{x,k} \in [\,u_{\mathrm{lo}},\,u_{\mathrm{hi}}\,]$ with
\begin{align}
\label{eq:u-interval}
u_{\mathrm{lo}} \doteq \frac{a\,L_{y,k}^- - p_{\max}}{c},\quad
u_{\mathrm{hi}} \doteq \frac{a\,L_{y,k}^- - p_{\min}}{c}.
\end{align}
Likewise, an energy envelope $E_{\min} \le E_{x,k+1} \le E_{\max}$ induces a ring constraint $r_{\min} \le |u_{x,k}| \le r_{\max}$ with 
\begin{align}
\label{eq:ring}
r_{\max}^2 &= \frac{2H}{g}\!\left(\frac{(L_{y,k}^-)^2}{2m^2H^2} - E_{\min}\right),\\
r_{\min}^2 &= \max\!\left\{0,\, \frac{2H}{g}\!\left(\frac{(L_{y,k}^-)^2}{2m^2H^2} - E_{\max}\right)\right\}.
\end{align}
Intersecting these with kinematic limits $u_{x,k}\in[u_{\min},u_{\max}]$ yields a feasible set that is the union of at most two intervals. The projection~\eqref{eq:proj-filter} is then obtained by snapping $u_{x,k}^{\text{nom}}$ to the nearest point in that set. The lateral direction is handled analogously with the separation constraint~\eqref{eq:sep-barrier} providing a half-space bound on $u_{y,k}$.


\section{Implementation on \textit{Digit}}
\label{sec:digit-impl}
We implement the ALIP-informed safe-RL framework on the \textit{Digit} humanoid in MuJoCo. This section details the estimator, controller, reward design, and RL training setup.

\subsection{Platform, Simulator, and Rates}
The RL policy runs at \(33\)~Hz (every \(30\)~ms), while the whole-body controller (WBC) operates at \(1\)~kHz. The pre-impact template state \(x_k^-\) is latched at foot-strike. The step duration is fixed at \(T = 0.35\)~s. The ALIP template states are obtained from Digit's proprietary estimator, which fuses IMU, joint kinematics, and contact information to estimate the horizontal CoM motion and the contact angular momentum components \((L_y,L_x)\).

\subsection{Observation Space}
The observations provided to the policy network are inspired by ALIP variables~\cite{castillo2023template} and defined as
\begin{equation}
    \mathbf{o} = 
    \begin{bmatrix}
        p_x,
        p_y,
        v_x^{avg},
        v_y^{avg},
        \tilde{v}_x^{avg},
        \tilde{v}_y^{avg},
        v_x^{des},
        v_y^{des}
    \end{bmatrix}^{\intercal},
\end{equation}
including the horizontal CoM position, average CoM velocity, velocity tracking error $\tilde{v}^{avg}\doteq v^{avg}-v^{des}$, and the desired horizontal CoM velocity.

\subsection{Action Space}
The action is a 3-dimensional vector specifying the swing foot location and torso pitch:
\begin{equation}
    \mathbf{a} = 
    \begin{bmatrix}
        u_{x,k} & u_{y,k} & \theta_{\text{pitch}}
    \end{bmatrix}^{\intercal},
\end{equation}
where $(u_{x,k},u_{y,k})$ are the desired horizontal touchdown locations of the swing foot \emph{relative to the CoM} at $t = T$ (CoM-to-foot), expressed in the current stance frame. In the Digit implementation, the WBC executes $(u_{x,k},u_{y,k})$ by using the estimated CoM to convert this command to a desired swing-foot pose relative to the base and tracking a minimum-jerk trajectory in $(x,y)$ (with a 5th-order B\'ezier height profile for clearance, $h_d = 0.15$~m). The torso pitch $\theta_{\text{pitch}}$ is included as a control variable to exploit torso motion for increased robustness.

\subsection{Whole-Body Control (1\,kHz)}
We employ a QP-based inverse-dynamics WBC~\cite{mistry2010inverse,herzog2016momentum,reher2021inverse} with tasks (in priority order): CoM height regulation to nominal \(H\), roll/yaw stabilization, torso pitch tracking \(\theta_{\text{pitch}}\), time-scaled swing-foot trajectories (min-jerk in \(x,y\), 5th-order B\'ezier in height), and posture/arm regularization. The QP enforces torque limits, friction cones, and holonomic contact constraints. Safety is \emph{not} enforced in the WBC; it is handled at the policy layer by the projection filter.

\subsection{Reward Composition}
We design the reward function to encourage the policy to track a desired horizontal velocity, to heuristically place the swing foot according to the commanded speed, and to minimize the discrepancy between the full-order model and the ALIP-based contact angular momentum. In particular,
\begin{equation}
    r = r_{\text{speed}} + r_{\text{foothold}} + r_{\text{ActionChange}} + r_{\text{pitch}} + r_{\text{ALIP}}.
\end{equation}
The unguided baseline uses this reward, whereas the DECBF-guided policy adds $r_{\text{safe}}$ from~\eqref{eq:rsafe}; the observation and action spaces, network, PPO setup, WBC, and training task are otherwise unchanged.

The speed reward encourages the robot to track the commanded velocity and is defined as 
$r_{\text{speed}} = 0.4 e^{-|v_x^{avg} - v_x^{des}|} + 0.4 e^{-|v_y^{avg} - v_y^{des}|}$,
where $(v_x^{avg}, v_y^{avg})$ denote the moving average velocity and $(v_x^{des}, v_y^{des})$ the desired velocity.

The foothold reward penalizes deviations from a reference swing foot placement computed from the desired walking speed.
It is given by 
$r_{\text{foothold}} = 0.1 e^{-10(u_{x,k} - p_x^{\text{ref}})^2} + 0.1 e^{-10(u_{y,k} - p_y^{\text{ref}})^2}$,
where $(p_x^{\text{ref}}, p_y^{\text{ref}})$ are the reference CoM-relative touchdown locations computed respectively as $(v^{des}_xT, \pm W+v^{des}_yT)$ with $W$ a user-defined ideal feet width.
To encourage smooth actions and reduce jitter, we include the action change reward 
$r_{\text{ActionChange}} = 0.1 e^{-10||\mathbf{a} - \mathbf{a}_{\text{old}}||^2}$,
with $\mathbf{a}$ the current action and $\mathbf{a}_{\text{old}}$ the previous one.
Large torso pitch angles are discouraged through the pitch reward 
$r_{\text{pitch}} = 0.1 e^{-|\theta_{\text{torso}}|}$,
where $\theta_{\text{torso}}$ denotes the torso pitch angle.
Finally, the ALIP reward enforces consistency with the Angular Linear Inverted Pendulum model. It is defined as 
$r_{\text{ALIP}} = 0.1 e^{-10 |L_c(t) - L_c^{\text{ALIP}}(t)|}$,
where $L_c(t)$ is the actual angular momentum about the contact point and $L_c^{\text{ALIP}}(t)$ the corresponding ALIP prediction.
\subsection{Learning Setup (PPO)}
The underlying neural network is a Recurrent Neural Network (RNN) with two hidden layers, each containing 256 units. We use the ReLU activation function for the hidden layers, while the output layer uses a sigmoid activation function followed by a scaling factor.
We use Proximal Policy Optimization (PPO) to train the network. A training episode terminates if the CoM height leaves the range $p_z \notin [0.8,\,2.0]$~m, indicating a fall or jump.

\subsection{ALIP-Informed Safety Bounds for \textit{Digit}}
We use stance-relative reach, lateral separation, and velocity (energy surrogate) bounds. The bounds used are summarized in Table~\ref{tab:safety}. When the safety filter is enabled, the foot-placement command \((u_{x,k},u_{y,k})\) is projected at each step onto the nearest command that satisfies the DECBF constraints (keeping the torso-pitch command unchanged).
We enforce the speed cap via the sagittal ALIP orbital energy (Sec.~\ref{sec:alip-rl}): using $L_y = mH v_x$, $E = -\frac{g}{2H}p_x^2 + \frac{1}{2}v_x^2$, so we set $E_{\max}=\frac{1}{2}v_{x,\max}^2 = 1.125$.

\begin{table}[t]
\centering
\vspace{2mm}
\caption{Safety bounds for \textit{Digit}}
\label{tab:safety}
\begin{tabular}{lcc}
\hline
Quantity & Value & Rationale \\
\hline
$p_{x,\min},p_{x,\max}$ & $(-0.70,\ 0.70)$ m & stance reach \\
$p_{y,\max}$ & $0.50$ m & lateral reach \\
$W_{\min}$ & $0.08$ m & no leg crossing \\
$v_{x,\max}$ & $1.5$ m/s & energy surrogate \\
$T$ & $0.35$ s & nominal step duration \\
\hline
\end{tabular}
\end{table}


\section{Results}
We evaluate our RL pipeline on the humanoid robot Digit in MuJoCo. A baseline (unguided) policy is trained with the task reward alone and compared against a DECBF-guided policy that additionally includes the safety-shaping reward $r_{\text{safe}}$ from~\eqref{eq:rsafe}. For each policy we evaluate performance with and without the projection-based safety filter from~\eqref{eq:proj-filter}. The test environment consists of flat terrain with periodic external disturbances:
\begin{equation}
    f_{ext}(t) = 
    \begin{cases}
        f_{\text{fixed}} =[300,300,0]^{\intercal}, &t \in [t^*_p + 3,\,t^*_p + 3.4] \\
        [0,0,0]^{\intercal}, &\text{otherwise}
    \end{cases}
    \label{eq: ext-dist}
\end{equation}
Here, $t_p^*$ is the most recent application time; the force repeats every 3~s for 0.4~s. The forward command steps from $0$ to $1.2$~m/s with zero lateral command.


\subsection{Leg crossing avoidance}
\begin{figure}[t]
\centering
\vspace{2mm}
\includegraphics[width=0.96\linewidth]{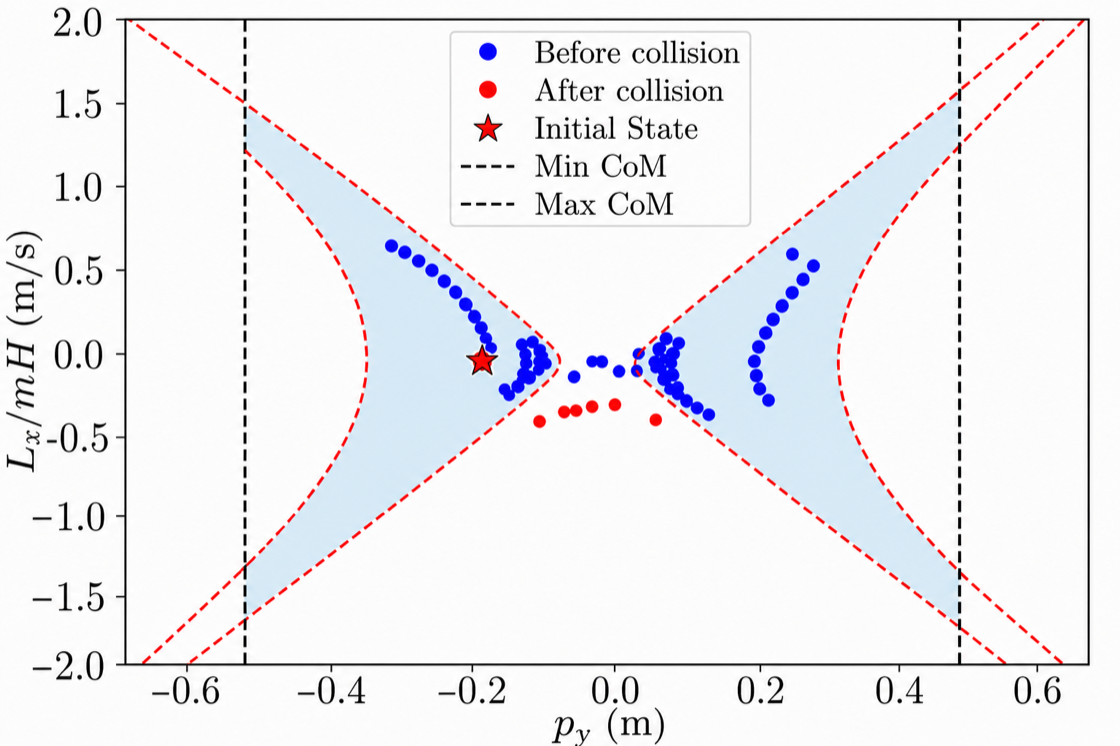}
\caption{Lateral-plane phase portrait showing the relationship between orbital energy and leg collision. During stable walking, the system remains in the safe region ($E<0$). After a disturbance drives the orbital energy positive (red states), the CoM crosses the stance foot and a leg collision occurs.}
\label{fig: PhasePortrait_Collision_P2}
\vspace{-1em}
\end{figure}

To illustrate the effect of the foot separation constraint, we evaluate the environment under external disturbances defined in (\ref{eq: ext-dist}) while tracking a desired velocity profile. Additionally, we illustrate how orbital energy can provide a warning of leg collision in this scenario. Fig.~\ref{fig: PhasePortrait_Collision_P2} shows a collision occurring after the orbital energy transitions from negative to positive.

\begin{figure*}[t]
    \centering
    \vspace{2mm}
    \includegraphics[width=0.99\linewidth]{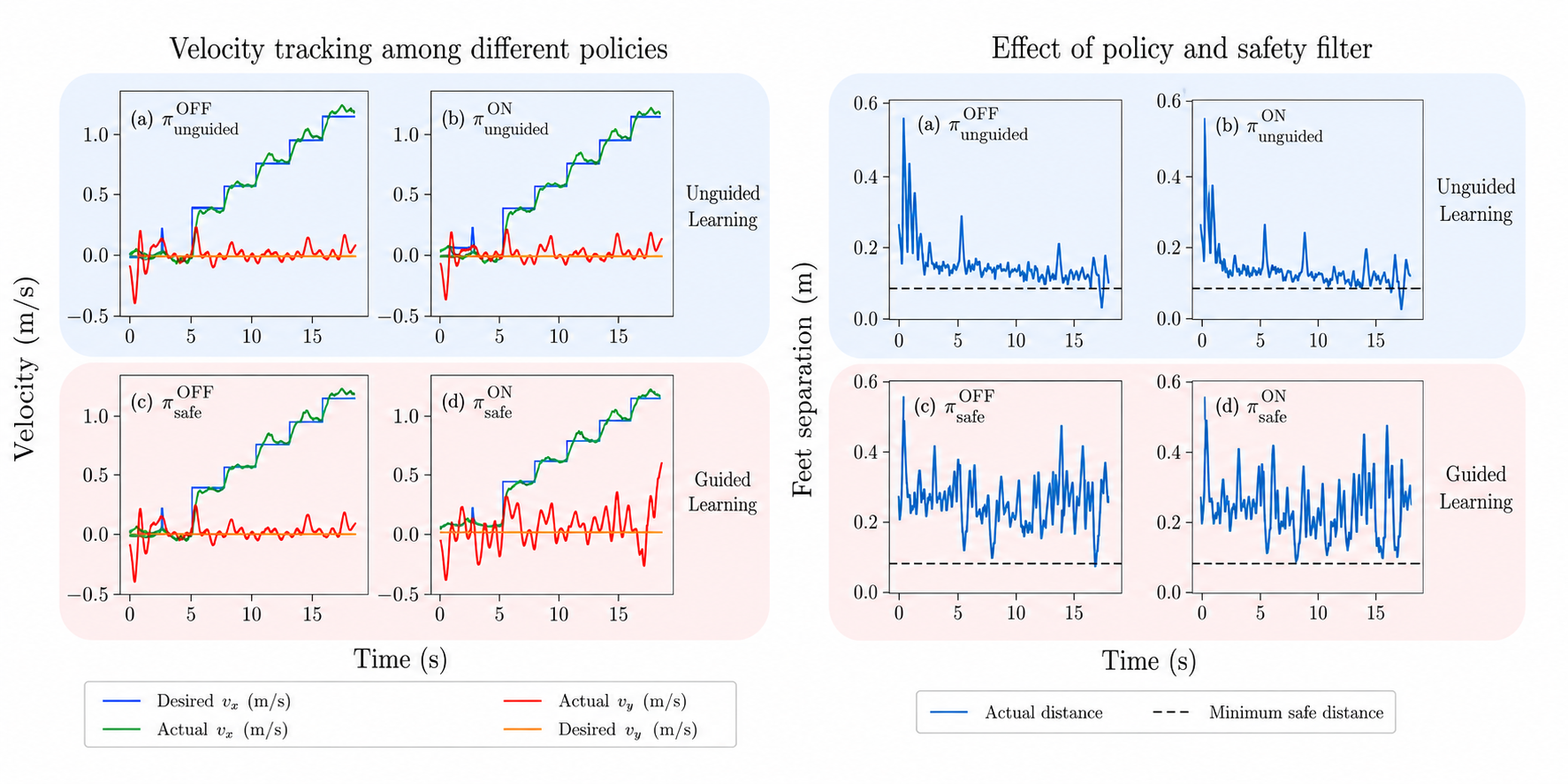}
    \vspace{-1em}
    \caption{Velocity tracking (left) and lateral foot distance (right) under periodic external disturbances. Left: desired vs.\ average CoM velocity for (a)~unguided, filter off; (b)~unguided, filter on; (c)~DECBF-guided, filter off; (d)~DECBF-guided, filter on. Right: lateral distance from stance foot to swing foot at touchdown compared against the safety bound $W_{\min}=0.08$~m. Subplots (a)--(d) follow the same ordering.}
    \label{fig:VelFootDistance}
\end{figure*}

We compare four policy variants, summarized in Fig.~\ref{fig:VelFootDistance}: unguided $\piunguidedoff$ (no safety), unguided with filter $\piunguidedon$, DECBF-guided $\pisafeoff$ (safety shaping only), and DECBF-guided with filter $\pisafeon$ (shaping + projection).



\subsubsection{Unguided baseline}
The baseline policy $\piunguidedoff$ tracks velocity commands effectively (Fig.~\ref{fig:VelFootDistance}a, left) but occasionally violates the lateral-separation bound, particularly at higher speeds (Fig.~\ref{fig:VelFootDistance}a, right). For most of the gait the feet remain close to the safety limit at $\pm 0.08$~m, indicating that the policy naturally learns narrow foot placement without explicit safety guidance.
Applying the safety filter without retraining ($\piunguidedon = \text{safety-filter}(\piunguidedoff)$) retains the overall tracking trend (Fig.~\ref{fig:VelFootDistance}b, left) and reduces violations (Fig.~\ref{fig:VelFootDistance}b, right). However, some violations persist because the filter enforces constraints at the planning level while the low-level whole-body controller may not perfectly track the projected commands. A WBC with built-in safety guarantees could mitigate this gap~\cite{paredes2024safe}.

\subsubsection{DECBF-guided policy}
With safety shaping alone ($\pisafeoff$), the policy continues to track the velocity command with visible transients (Fig.~\ref{fig:VelFootDistance}c, left) while violations become far less frequent and the average foot separation stays well away from the bound (Fig.~\ref{fig:VelFootDistance}c, right). This indicates that the DECBF reward successfully steers the policy toward safer foot placements during training.
Adding the projection filter ($\pisafeon$) eliminates measured violations in this trial (Fig.~\ref{fig:VelFootDistance}d) but introduces larger lateral-velocity oscillations and settling time, indicating a safety--tracking tradeoff. Notably, the foot separation under this policy is larger than the unguided case, but this is expected: the safety constraint only imposes a \emph{minimum} separation, so wider steps are permissible and often preferred for robustness.

\subsubsection{Safety-violation metric}
To quantify these observations, we define a normalized violation metric $\mathcal{M}_{\pi}$ based on the lateral-separation barrier $B_{\text{sep}}$ in~\eqref{eq:sep-barrier}:
\begin{align}
    m_{\pi} &= \sum_{k=1}^N \max\!\left(0,\,-B_{\text{sep},k}\right)\Big|_{\pi},\quad
    \mathcal{M}_{\pi} = \frac{m_{\pi}}{ \max_{\pi \in \Pi}\{ m_{\pi}\}}\nonumber
\end{align}
where $\mathcal{M}=1$ corresponds to the worst tested policy. With the filter off, the unguided policy scores $\mathcal{M}=1.0$ while the DECBF-guided policy reduces to $0.13$—an $87\%$ reduction from safety shaping alone. With the filter on, the DECBF-guided policy achieves zero measured violations, and the unguided policy's metric drops to $0.23$. Because $m_\pi$ accumulates the negative barrier depth at each touchdown, $\mathcal{M}_\pi$ reflects both the occurrence and magnitude of lateral-separation violations. These results indicate that safety shaping and projection are complementary: shaping reduces unsafe actions during training, while the filter enforces the template-level constraint at execution. These are trial-level measurements, not full-order guarantees.

\begin{figure}[t]
\centering
\vspace{2mm}
\includegraphics[width=0.92\linewidth]{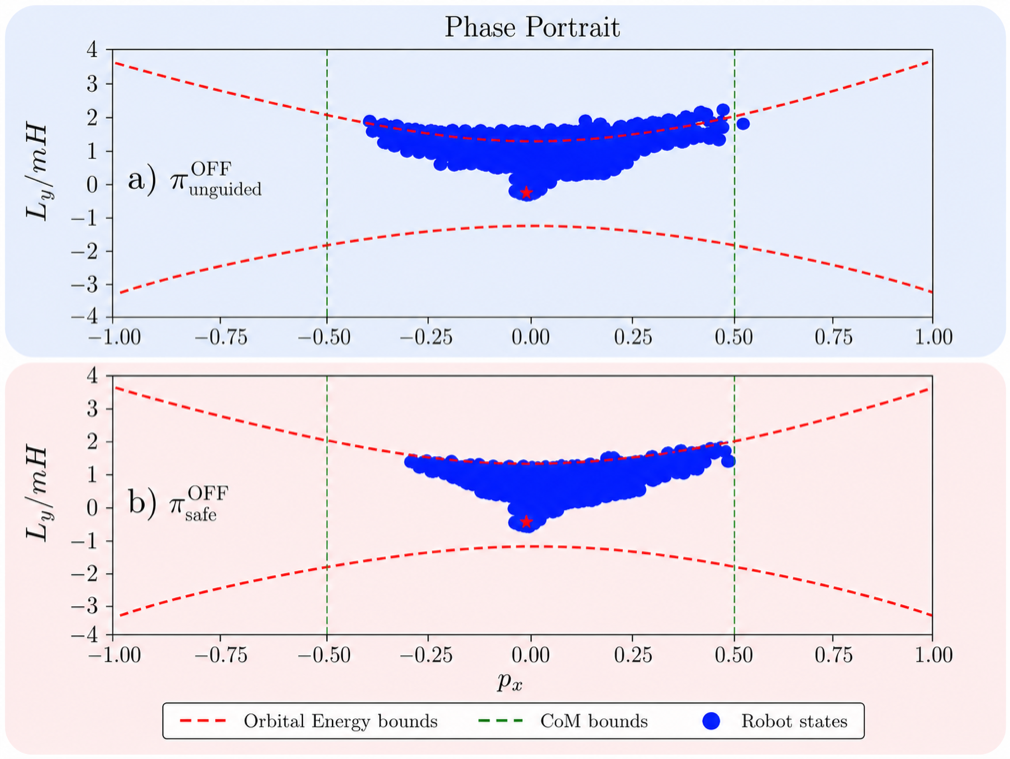}
\caption{Sagittal phase portrait under periodic disturbances~\eqref{eq: ext-dist}. (a)~The unguided policy violates the orbital-energy and CoM-extension bounds. (b)~The DECBF-guided policy remains within the safe region. The red star marks the initial state.}
\label{fig: P1-Safe-Results}
\vspace{-1em}
\end{figure}

\subsection{Maximum CoM extension via DECBF}

We additionally apply the sagittal orbital-energy and CoM-reach DECBFs from Sec.~\ref{sec:alip-safe} to regulate walking speed, using $p_x \in [-0.7,\,0.7]$~m and $v_x \le 1.5$~m/s. Fig.~\ref{fig: P1-Safe-Results} shows the sagittal phase portrait under external disturbances. The unguided trajectory exceeds the prescribed orbital-energy and CoM-reach bounds (a), whereas the DECBF-guided trajectory remains within the plotted template-level region (b) in the reported trial.

\subsection{Discussion}
Shaping changes nominal footholds, whereas projection modifies a selected command only at execution. Because all variants share the policy architecture, PPO setup, and WBC, paired comparisons isolate these effects; the guided policy's larger separation suggests that shaping moves footholds away from the boundary before projection.
Across both experiments, two observations stand out. First, safety shaping alone (filter off) substantially reduces violations but cannot guarantee them, since the policy may still select borderline actions that the stochastic optimization does not fully penalize. The projection filter reduces this gap by enforcing the template-level constraint at execution time. Second, the remaining violations under the filtered unguided policy show that model mismatch and WBC tracking error separate command feasibility from full-order safety. The larger lateral-velocity transients further indicate a safety--tracking tradeoff.
From a computational standpoint, the closed-form projection requires only scalar comparisons and clamp operations over at most two intervals per planning step.

\section{Conclusions}
We presented a model-informed safe RL framework that anchors policy learning to the ALIP step-to-step map and enforces safety at the impact horizon via Discrete Exponential Control Barrier Functions (DECBFs). The DECBF condition is used in two complementary ways: as a training-time shaping reward that steers exploration toward safe foot placements, and as an execution-time projection filter that minimally adjusts the policy output to satisfy template-level constraints.
In the reported Digit MuJoCo trial under periodic external disturbances, the DECBF-guided policy with the safety filter has no measured lateral-separation violations and keeps the sagittal orbital energy within prescribed bounds, while larger lateral-velocity transients reveal a safety--tracking tradeoff. The unguided baseline, by contrast, exhibits more frequent constraint violations and leg-crossing events.
Several limitations remain. The DECBF operates on the reduced-order ALIP model; residual model mismatch and imperfect low-level tracking can still produce minor full-order violations. The current evaluation is limited to flat terrain with a fixed step duration. Incorporating a continuous-time Exponential CBF within the whole-body controller~\cite{paredes2024safe} would provide an additional layer of full-order safety. Future work will also address variable step timing, uneven terrain, and hardware deployment on the physical Digit platform.




\bibliography{IEEEabrv,IEEEexample}
\bibliographystyle{IEEEtran}
\end{document}